\RequirePackage{fix-cm}
\documentclass[smallextended]{svjour3}       % onecolumn (second format)
\smartqed  % flush right qed marks, e.g. at end of proof
\usepackage{graphicx}
\usepackage[utf8]{inputenc}
\usepackage{amsmath}
\usepackage{amssymb}
\usepackage{natbib}
\setcitestyle{aysep={}}
\usepackage{hyperref}

\journalname{Extremes}

\begin{document}

\title{BlackBox: Generalizable Reconstruction of Extremal Values from Incomplete Spatio-Temporal Data%\thanks{Grants or other notes
}

\titlerunning{BlackBox: Generalizable Reconstruction of Extremal Values}        % if too long for running head

\author{Tomislav Ivek \and Domagoj Vlah}
\institute{Tomislav Ivek \at
Institut za fiziku, Bijenička 46, HR-10000 Zagreb, Croatia\\
\email{tivek@ifs.hr}
\and
Domagoj Vlah (corresponding author)\at
University of Zagreb, Faculty of Electrical Engineering and Computing, Department of Applied Mathematics, Unska 3, HR-10000 Zagreb, Croatia\\
\email{domagoj.vlah@fer.hr}
}

\date{Received: 30 April 2020 / Accepted: date}

\maketitle

\begin{abstract}
We describe our submission to the Extreme Value Analysis 2019 Data Challenge in which teams were asked to predict extremes of sea surface temperature anomaly within spatio-temporal regions of missing data. We present a computational framework which reconstructs missing data using convolutional deep neural networks. Conditioned on incomplete data, we employ autoencoder-like models as multivariate conditional distributions from which possible reconstructions of the complete dataset are sampled using imputed noise. In order to mitigate bias introduced by any one particular model, a prediction ensemble is constructed to create the final distribution of extremal values. Our method does not rely on expert knowledge in order to accurately reproduce dynamic features of a complex oceanographic system with minimal assumptions. The obtained results promise reusability and generalization to other domains.

\end{abstract}

\section{Introduction}
\label{intro}

The EVA 2019 Data Challenge posited a problem to predict extremes of the Red Sea surface temperature anomaly within spatio-temporal regions of missing data \citep{Huser2020}. Daily temperature anomaly values were provided for contestants spanning over $31$ years and covering the geographical area of the Red Sea. For each day, temperature anomaly values were given at fixed spatial points on a regular geographical grid. About $31.6\%$ of data was deliberately removed from the dataset. Regions of the missing data were approximately contiguous with irregular boundaries, relatively large, at least one calendar month in duration, and present for every calendar day in the provided dataset. The exact process of data removal was not disclosed to contestants. The goal was to predict the distribution of extremes of temperature anomaly on a number of specified space-time cylindrical regions ($50$km in radius and 7 days in length), chosen in the most difficult part of the dataset which had $60\%$ percent of data missing for any day. The quality of predicted extremes was evaluated using the threshold-weighted continuous ranked probability score averaged over all prediction regions $\overline{\mathrm{twCRPS}}$ \citep{Huser2020}.

Recently there has been an increase in adoption of deep neural network models in various areas of research and technology which feature high-dimensional interdependent data including time series and imaging. Inspired by some of these advances \citep{Asadi2019,Li2019,Schlemper2017_2} we used state-of-the-art neural network techniques primarily from image processing in order to complement existing extreme value theory approaches for spatial extremes \citep{davison2012,davison2015,davison2019}. Deep learning techniques typically do not depend on expert domain knowledge and generalize well to data from different domains. However, they do rely on large quantities of data being available for model training. Furthermore, currently the engineering of training and deploying neural networks is ahead of their precise theoretical treatment, especially regarding error bounds, model optimization in the sense of minimization of loss function with potentially billions of parameters and associated rate of convergence, optimal model complexity and size, and the related speed of model inference.

In the absence of expert domain knowledge, to predict the distribution of extremes of Red Sea surface temperature anomaly within regions of withheld data we attempt to reconstruct the missing information. We introduce additional damage, i.e., we remove even more data from the original dataset in order to teach autoencoder-like models based on convolutional deep neural networks how to repair or in-paint the missing data based on the remaining information. Then, we evaluate trained models on originally provided data in order to create stochastic plausible reconstructions of temperature anomaly within regions of missing data. The extremal values within regions of interest are then trivially calculated for each stochastically sampled reconstruction which finally allows us to create their distribution. We discuss details of our implementation, possible extensions to the technique and its generalizability to different problem domains.

\section{Overview of related work}
\label{prev_work}
We first review some of the recently published techniques based on neural networks relevant to missing data reconstruction. Neural networks are parameterizable approximators based on multiple compositions of affine and nonlinear functions which are fitted to, or trained on, some desired dataset. They have long been used to recover data gaps in time series, including geophysical datasets \citep{Rossiev2002,Lee2015}. Conventionally, raw data is first reduced in dimensionality and mapped onto a small-dimension manifold, also called latent or hidden space, which aims to capture salient features underlying the modeled phenomena. In case of time series, latent vectors are then reconstructed at missing time stamps by a neural network predicting the next step based on history. Finally, the predicted reconstruction is projected back to original data space. Various model architectures are researched here and used in production, most common being simple fully connected networks \citep{Rossiev2002}, recurrent neural networks \citep{Che2018}, which are either fully-connected for tabular data or combined with convolutional neural networks where spatial or temporal proximity is important \citep{Asadi2019}. Particularly interesting is the recently introduced BRITS architecture \citep{Cao2018}, which uses a novel bidirectional recurrent component based on learned feature correlation and temporal decay in order to impute data. While this technique seems most promising for tabular data with measured or engineered features of interest, it needs to be adapted in order to handle image-like data on a large spatial grid.

Since the number of parameters of neural layers grows proportionally to dimensionality of their input, fully connected neural networks where each element of input influences the whole of output are often deemed intractable for large image-like inputs. Moreover, in computer vision and image processing it is commonly desirable that algorithms operate independently of feature position, e.g., a face detection network should correctly identify human faces regardless of their position in a photograph. With this in mind, convolutional neural networks \citep{Zhang1988} have become the prevalent choice for modeling ordered data on space-like grids. Such an architecture again comprises layers, each with a small common neural network, or ``kernel'', which slides over the whole input. Output of a convolutional layer can be regarded as a spatial map of detected learnable features which grows semantically richer with every consecutive layer. Thanks to their shared-weight architecture and local connectivity \citep{Behnke2003}, convolutional neural networks train well on smaller datasets, generalize to unseen data examples, and are used with great success in various classification and prediction tasks \citep{Krizhevsky2012,Schlemper2017_2}.

Recently an innovative convolutional framework was proposed for missing data reconstruction called MisGAN \citep{Li2019}. It is particularly suited for high-dimensional data with underlying spatial correlations. MisGAN uses multiple generative adversarial networks \citep{Goodfellow2014} where the so-called generators create increasingly more convincing fake samples as well as their missing data masks, while the discriminators attempt to discern them from real samples. This complex scheme simultaneously learns the distribution of missing data, or ``masks'', as well as the conditional distribution of data predicated on masks. It finally constructs a probabilistic imputer model which repairs the data by sampling from the learned distribution given some known data and its mask.

MisGAN provides state-of-the-art reconstruction results on several standard datasets and appears to be particularly suited to the task at hand. However, in our preliminary experiments to apply MisGAN on Red Sea temperature anomalies it was difficult to achieve convergence. Generative adversarial networks are notoriously difficult to train as they are a minmax problem where the optimal state is a saddle point with a local minimum in generator network cost and local maximum of the discriminator network cost \citep{Wei2018,Le2017}. In our admittedly limited tests, MisGAN tended to diverge and create patterned artifacts. Furthermore, the training itself took prohibitive amounts of time while it also appeared the provided amount of data was not sufficient.

Due to these issues, for the particular problem of Red Sea temperature anomalies we opt to step back from generative adversarial networks and construct a simpler framework based on autoencoder networks \citep{Kramer1991,Goodfellow2016}. Still, MisGAN provides us with certain valuable tools and avenues to explore. We adopt its use of noise-imputed samples in order to treat trained models as conditional distributions. Samples of temperature anomaly generated from our models are conditioned on the known temperature anomaly data for a given day. In that sense their distribution is conditioned on known data. Also, we use a simpler version of convolutional networks to exploit spatial coherence and short-range temporal correlations. We forego long-term dependencies and leave them for future consideration.

\section{Model and methodology}
\label{model_and_meth}
We aim to construct one or more models which would take as input incomplete, damaged temperature anomaly data and attempt its best guess to reconstruct or predict the original complete data. In the following section we introduce separate ``ingredients'' which come together to form a powerful framework for probabilistic data repair. First, we present a general notion of models which map from incomplete to complete data. Then, we introduce the concept of autoencoder neural networks and modify their fitting procedure to take into account missing data. We also discuss the concept of spatial coherence and the benefits of using convolutional models. Our model training protocol is described. The obtained fitted models take as input incomplete temperature anomaly data and infer repaired data from which temperature anomaly extremes are trivially computed. In the end, we use model ensembling to improve predictions of extremes distributions. 

\subsection{Trivial case: target data is complete and available for model training}

Starting with the most simple case, let us for now disregard time dependence of temperature anomaly data and consider each day as a separate multi-dimensional point. Suppose we have the desired input-output relations $\left\{(x^{(n)}_\mathrm{orig}, y^{(n)}_\mathrm{complete}); n \in \{1,...,T\}\right\}$, where $T$ is the number of days included in the dataset, $x^{(n)}_\mathrm{orig} \in \mathbb{R}^{W \times H}$ is a matrix representing originally incomplete or damaged data provided of weight $W$ and height $H$ in the problem statement and $y^{(n)}_\mathrm{complete} \in \mathbb{R}^{W \times H}$ the matrix representing ideal, undamaged data for each day $n$. If $\theta$ designates all the parameters of some model attempting to summarize these relations, their optimal value to reconstruct the missing data based on $x^{(n)}_\mathrm{orig}$ as input can be obtained by minimizing the loss function
\begin{equation}
\mathcal{L} = \frac{1}{T}\sum_{n=1}^{T} \ell\left(y^{(n)}_\mathrm{complete}; o^{(n)}(\theta)\right)
\label{cost}
\end{equation}
where $o^{(n)}(\theta) \in \mathbb{R}^{W \times H}$ is the model output predicated on parameters $\theta$ for model input $x^{(n)}_\mathrm{orig}$, for each day $n$, and $\ell$ is a suitable distance function between targets and corresponding model outputs.

In principle, the posited problem could be solved by taking $\mathcal{L}$ to be the threshold-weighted continuous ranked probability score ($\mathrm{twCRPS}$) averaged over all space-time validation points $s,t$ specified by the Data Challenge, where
\begin{equation}
	\mathrm{twCRPS}(\hat{F}_{s,t},u_{s,t})=\int_{-\infty}^{\infty}\left\{\hat{F}_{s,t}(u)-\mathbb{I}(u_{s,t})\leq u)\right\}^2 w(u)\,\mathrm{d}u.
	\label{twCRPS}
\end{equation}
$\hat{F}_{s,t}$ denotes the distributions of extremes of predicted temperature anomaly $o^{(n)}$, and $u_{s,t}$ is the observed extremes of temperature anomaly $y_\textrm{complete}^{(n)}$, $\mathbb{I}(\cdot)$ is the indicator function, $w(x)=\Phi\{(x-1.5)/0.4\}$, and $\Phi(\cdot)$ is the standard normal distribution. Both predicted and observed extremes are evaluated over the spatio-temporal cylinder around spatial location $s$ and day $t$. For more details see \citep{Huser2020}. Inconveniently, the complete data $y^{(n)}_\mathrm{complete}$ is unavailable by the very nature of the problem we wish to solve so the average $\mathrm{twCRPS}$ cannot be taken as an optimization target. Therefore, a working solution will necessarily grow more complex as we need to find an adequate proxy cost function to minimize.

\subsection{Extracting salient information by introducing additional damage}
The provided daily temperature anomaly data on a geographical grid can be regarded as a time series of raster images represented by real matrices with $W$ columns in width and $H$ rows in height where each matrix element corresponds to the value of temperature anomaly at a certain geolocation for day $n$. We introduce its masking matrix $m^{(n)}_\mathrm{orig} \in \{0,1\}^{W\times H}$ which describes the extent of damage present in $x^{(n)}_\mathrm{orig}$. It carries binary information for each spatial location: 1 encodes that a value is observed and available and 0 designates unobserved values e.g., due to damage or the location itself not being present in the dataset \citep{Cao2018,Li2019}.

Let us now introduce additional data loss to the already damaged original data. We describe the total damage i.e., newly introduced damage together with the originally missing data, in the form of a new binary masking matrix $m^{\prime(n)} \in \{0,1\}^{W\times H}$. The additionally damaged data matrix then becomes
\begin{equation}
x^{\prime(n)} = x_{\mathrm{orig}}^{(n)} \odot m^{\prime(n)},
\end{equation}
where $\odot$ denotes element-wise multiplication of two matrices. We discuss generating $m^{\prime(n)}$ later in the text.

This setup in principle allows us to train a model which maps data with additional damage $x^{\prime(n)}$ onto the data with original amount of damage as the target, $y^{\prime(n)} = x^{(n)}_\mathrm{orig}$. Our main idea is that by training to repair the additional damage, a sufficiently powerful convolutional model could learn to extract features underpinning the data manifold which are robust to damage. Then, when such a trained model is applied to the original data, we hypothesize it will be able to reconstruct missing data in an adequate manner.

\subsection{Weighted distance function}
The training goal also needs to codify that the relevant cost is evaluated only where data is provided by the inherently damaged training target. For the distance function $\ell$ in (\ref{cost}) we substitute weighted $L^1$ or $L^2$ distances evaluated exclusively on the original masks $m^{(n)}_\mathrm{orig}$ so that missing or unobserved data is ignored:
\begin{equation}
\ell\left(y^{(n)}, m^{(n)}_\mathrm{orig} ; o^{(n)}\right) = \frac{\left\Vert \left(y^{(n)} - o^{(n)}\right)\odot m^{(n)}_\mathrm{orig}\right\Vert}{\left\Vert m^{(n)}_\mathrm{orig}\right\Vert_{1,1}}.
\label{distance}
\end{equation}
Here $\left\Vert \cdot \right\Vert$ in the numerator denotes either $L^1$ or $L^2$ vector norm i.e., $L_{1,1}$ and $L_{2,2}$ matrix norms, which are defined as
\[
\Vert A\Vert_{r,r} = \left(\sum\limits_{i=1}^W\sum\limits_{j=1}^H |a_{ij}|^r\right)^{1/r} ,
\]
where $A=(a_{ij})\in\mathbb{R}^{W\times H}$, for $r=1,2$.
In the denominator $\left\Vert \cdot \right\Vert_{1,1}$ is strictly the $L_{1,1}$ matrix norm i.e., the sum of absolute values of all mask elements effectively counting the number of observed temperature anomaly values in a particular day. Note that the computation of $o^{(n)}$ requires additionally damaged data $x^{\prime(n)} = x^{(n)}_{\mathrm{orig}} \odot m^{\prime,(n)}$, but the distance $\ell$ being minimized in (\ref{distance}) utilizes the original mask $m^{(n)}_\mathrm{orig}$.

With such modifications in place, the training procedure will create models that reproduce known data but are not punished when ``speculating'' on regions of missing data.

\subsection{Sampling from trained models as multivariate probability distributions}
We found it most fruitful to introduce both a level of stochasticity to model training input, as well as to the evaluation input fed into trained models. Specifically, we impute noise wherever masks indicate missing data:
\begin{align}
x^{\prime(n)}_\mathrm{noise} &= \left[x^{(n)} \odot m^{\prime(n)} +
 \mathcal{N}^{(n)} \odot \left(1 - m^{\prime(n)}\right)\right] \odot m_\mathrm{master},\\
x^{(n)}_\mathrm{orig,noise} &= \left[x^{(n)} \odot m^{(n)}_\mathrm{orig} +
 \mathcal{N}^{(n)} \odot \left(1 - m^{(n)}_\mathrm{orig}\right)\right] \odot m_\mathrm{master},
\end{align}
where $\mathcal{N}^{(n)}\in \mathbb{R}^{W\times H}$ is noise sampled independently for each spatio-temporal location and $m_\mathrm{master}\in \{0,1\}^{W\times H}$ is the master mask with 1 at every valid spatial location and 0 otherwise which effectively removes any noise spilling outside of the valid geographical Red Sea region.

Setting the distribution of imputed noise $\mathcal{N}$ equal to the expected distribution of missing data allows the model to learn expected ranges of valid data within the damaged regions. Moreover, imputed noise in evaluation input allows us to stochastically sample from the trained model, effectively using it as a multivariate conditional probability distribution. Such a sampling procedure has the most desirable property that the dimensionality of noise is proportional to the amount of data loss: the more information the model receives as input, the less variation it creates at its output \citep{Li2019}.

\subsection{Convolutional autoencoder architecture}
Finally, we describe a flexible family of parameterizable functions with sufficient capacity to learn patterns inherent to the provided incomplete matrix data and provide reasonable reconstructions of the complete pristine data.

The models we employ are sequential convolutional neural networks similar to autoencoder neural networks \citep{Kramer1991,Goodfellow2016}. In short, autoencoders are models which are fitted to approximate the identity function by preserving only the most relevant features of the data. They could be regarded as a composition of two fittable functions. The first function, the encoder, maps the input to a usually lower-dimensional ``latent'' space, effectively compressing the data. The result is mapped by the second function, the decoder, back to the original space. The fitting procedure ensures that autoencoder output is close to its input by minimizing some chosen loss function. The encoder-decoder architecture is robust to input noise and in principle may allow it to generalize to data not seen during the fitting procedure. In our case, we do not approximate an identity function but wish to reconstruct missing data. With this goal we change the autoencoder fitting procedure to use masks as described in the previous sections.

\begin{figure}
\includegraphics[clip,width=1.0\linewidth]{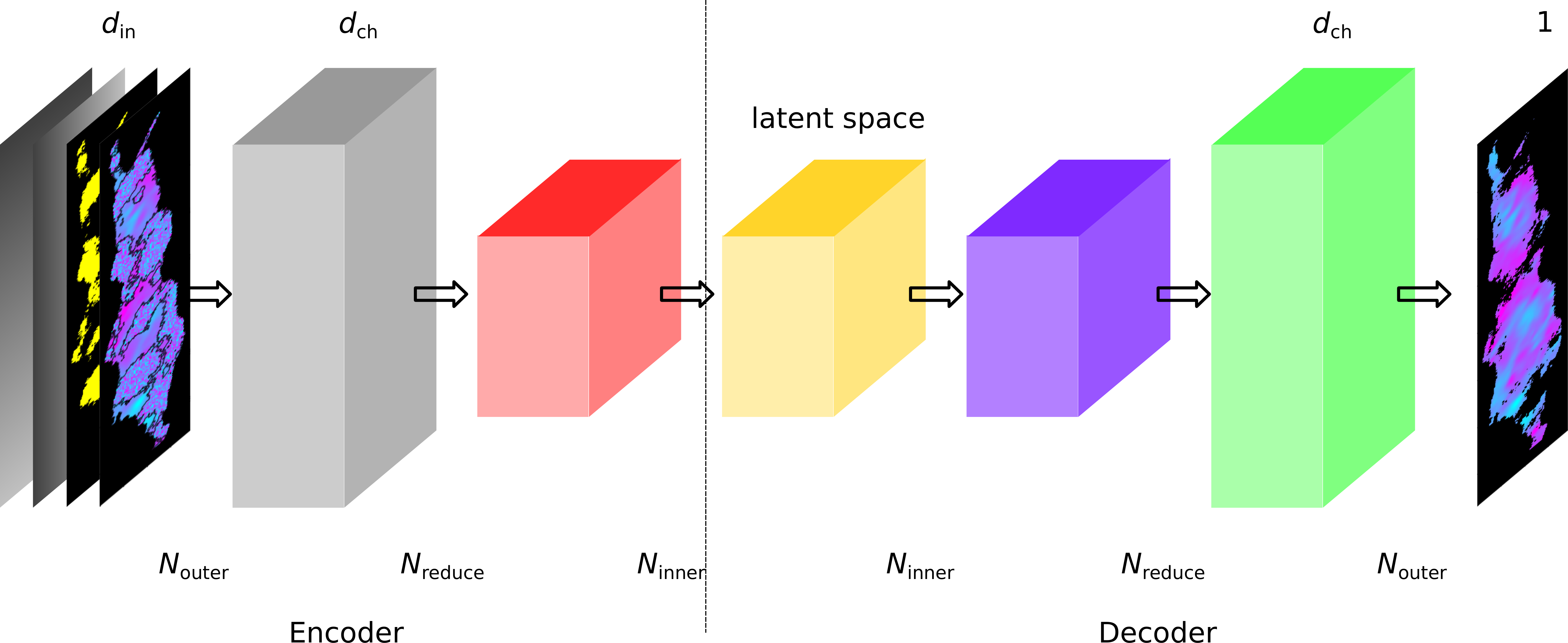}
\caption{Architecture of the convolutional autoencoder model. At encoder input a tensor with $d_\mathrm{in}$ channels is given containing several days worth of data (training data depicted, see text). The first block of $N_\mathrm{outer}$ convolutional layers increase the number of channels to $d_\mathrm{ch}$. Then, $N_\mathrm{reduce}$ layers reduce the spatial extents of the tensor. Following those, $N_\mathrm{inner}$ convolutional layers map the tensor into latent space. The decoder structure is approximately symmetric to the encoder. The model outputs a single-channel tensor as prediction.}
\label{fig:CNN}
\end{figure}

Encoders and decoders could be regarded as compositions of functions, so-called layers. Each layer is in principle a composition of a fittable affine mapping and an element-wise nonlinear function. In order to detect and utilize any spatial correlations inherent to our data, for the affine mapping we use 2D matrix convolutional operations \citep{Krizhevsky2012,Schlemper2017_2}. Figure \ref{fig:CNN} shows the architecture of our convolutional model. The domain of the first layer of the network is $\mathbb{R}^{d_\mathrm{in} \times W \times H}$. So, on the input we have a total of $d_\mathrm{in}$ matrices of size $W \times H$ which may comprise e.g., temperature anomalies for a number of consecutive days and their masks. The output of this first layer is a tensor from space $\mathbb{R}^{d_\mathrm{ch} \times W \times H}$, where $d_\mathrm{ch}$ is conventionally called the number of channels. Subsequent $N_\mathrm{outer}$ encoder layers map from $\mathbb{R}^{d_\mathrm{ch} \times W \times H}$ to $\mathbb{R}^{d_\mathrm{ch} \times W \times H}$ i.e., they do not change the spatial size or number of channels. However, the next $N_\mathrm{reduce}$ layers each reduce the spatial extents of data by a factor of 2 using convolutions of stride 2, more specifically every reducing layer $i = 1,\ldots, N_\mathrm{reduce}$ maps from $\mathbb{R}^{d_\mathrm{ch} \times \left(W/2^{i-1}\right) \times \left(H/2^{i-1}\right)}$ to $\mathbb{R}^{d_\mathrm{ch} \times \left(W/2^i\right) \times \left(H/2^i\right)}$. Finally, the last set of $N_\mathrm{inner}$ encoder layers preserves tensor dimensions and maps to latent space $\mathbb{R}^{d_\mathrm{ch} \times \left(W/2^{N_\mathrm{reduce}}\right) \times \left(H/2^{N_\mathrm{reduce}}\right)}$. The decoder closely follows this architecture in reverse using transposed convolutions instead of convolutions and ends with an additional single layer which contains only a transposed convolution and outputs a so-called one-channel tensor $\mathbb{R}^{W \times H}$ corresponding to the reconstructed temperature anomaly.

We can use a block of consecutive days as input by setting $d_\mathrm{in} > 1$. Inspecting several days of data to reconstruct a single day's temperature anomaly exploits short-term temporal correlations inherent to oceanographic data and teaches the model to also use rates of change instead of separate ``snapshots'' at a single point in time.

One typical advantage of convolutional neural networks, translational invariance, can become a hindrance when data might depend on absolute position in space. In our case, it may very well be that Red Sea temperature anomalies consistently differ depending on geographical location. Inspired by recent developments in natural language processing \citep{Vaswani2017}, we optionally concatenate two additional channels to the input of the first convolutional layer which provide information about absolute geographical latitude and longitude of each point as horizontal and vertical linear sweep of real numbers between $-1$ and $1$. In this way a model might find it advantageous to use this so-called positional encoding for greater selectivity and more precise predictions.

\subsection{Training protocol and implementation}

Our model, training and evaluation code is available online at \url{https://github.com/BlackBox-EVA2019/BlackBox}.

\paragraph{Training dataset and validation dataset.}
In order to train our models we separate the provided data $(x^{(n)}_\mathrm{orig},m^{(n)}_\mathrm{orig})$, per day basis, into two datasets, named (as customary in machine learning model) the training and the validation dataset. The training dataset is used to train models. The validation dataset is used during model development to measure the generalization capability of models on unseen data. Finally, the test set, data witheld from contest participants, is not available to us and was not used for model training purposes or optimization of model architecture.

Only after the models are trained and evaluated, we calculate the $\overline{\mathrm{twCRPS}}$ score of our final predictions using the observed distribution of extremal values which was provided by the contest organizers after the final ranking was determined.

Notice here the important difference between our naming scheme and the one in the Data Challenge problem statement \citep{Huser2020}. In the problem statement, the whole available data $x_\mathrm{orig}^{(n)}$ where $m^{(n)}_\mathrm{orig}=1$ is called the training dataset, and the subset of rest of the data hidden to contestants where $m^{(n)}_\mathrm{orig}=0$ but $m^{(n)}_\mathrm{master}=1$ is called the validation dataset.

Not all days in the Red Sea dataset have the same amount of non-missing data. The available time window spans over a total of $31$ years, but for the first $22$ years 20\% of data per day is missing, and for the last $9$ years 60\% of data per day is missing. Incidentally, the latter 9 years with more missing data also represents the period of accelerated climate change. The validation set has to be representative of all data, so we choose it as $5$ continuous years where $22/31$ parts are from the first $22$ years and $9/31$ parts are from the last $9$ years i.e., data belonging from $6736$th to $8560$th day. The remaining data is used for the training set. In this way both our training and validation datasets consist of large contiguous blocks of time, have the same distribution of missing data percentage, and potentially both contain data from the period of accelerated climate change.

\paragraph{Generating additional masks.}
Regarding the model training process, as already stated in Section \ref{model_and_meth}, we introduce further damage to the data for training input. A nontrivial question is how much additional data we should mask. We consider the natural choice to mask the same percentage of data for the process of training as would be masked during model inference when generating predictions of unknown test data. In the last $9$ years of data, where we are tasked to generate predictions, $60\%$ of data is masked. Therefore we need to create masks $m^{\prime(n)}$ such that approximately $60\%$ percent of data is removed for each day in training and validation datasets. Notice that the total data loss by masking $m^{\prime(n)}$ is $68\%$ for the first $22$ years of data, and even $84\%$ for the last $9$ years!

A further issue is how to actually generate masks $m^{\prime(n)}$. Our limited  experiments with training generative adversarial networks \citep{Li2019} to create convincing masks did not produce desired results. Therefore it is important to devise a method to create adequate masks $m^{\prime(n)}$. Except noting that $m^{(n)}_\mathrm{orig}$ changes once only every calendar month so our generated masks need to do the same, the exact mechanism by which $m^{(n)}_\mathrm{orig}$ were generated is unknown to us. We developed a stochastic diffusion algorithm that generates random masks similar in appearance to those provided by the problem statement. Our algorithm parameters needed to be manually tweaked until generated masks $m^{\prime,(n)}$ appeared visually indistinguishable from $m^{(n)}_\mathrm{orig}$, which is certainly one obvious drawback of our method.

\paragraph{Noise imputation.}
The imputed noise is sampled from a Gaussian distribution closely mimicking the marginal distribution of all provided temperature anomaly measurements, parameterized by $\mu=-0.0365\,^\circ\text{C}$ and $\sigma=0.683\,^\circ\text{C}$.

\paragraph{Reducing the data footprint.}
The original anomaly data is spatially large and taxes the capacity of current GPU architectures, both in sense of used memory and computation time. We used two complementary approaches to successfully reduce the GPU footprint.

Using only a small number of padding rows and columns at the edges, the original temperature anomaly data fits into a spatial matrix of $W \times H = 256 \times 384$. Notice that $256$ is a power of $2$ and $384$ is $3$ times a power of $2$, which turns out to be essential for efficient computation with convolutions. However, note that the Red Sea is elongated but geographically laying in the direction NNW-SSE so in our rectangular image representation most image elements correspond to land masses which carry no relevant data.

In order to increase the density of usable data we skew every other image row together with remaining rows below it towards west, beginning from the top to the bottom of image. In this way the whole Red See can fit in an image $W \times H = 96 \times 384$, while still preserving spatial coherence. At this point the data size is reduced but still leaves an unacceptably large GPU footprint.

Further, notice that our skew operation results with spatial extents of $96 \times 384$ which are divisible by $3$. We can conveniently down-sample the data by taking the average anomaly value over $3 \times 3$ spatial cells which brings the data size down by almost an order of magnitude, a quite substantial amount. We get the final anomaly matrix resolution of $W \times H = 32 \times 128$. Both numbers are factors of $2$ which is suitable for efficient GPU computations. Our models are both trained and inferred in this lower resolution.

To generate a prediction from a model trained on such data, after inference with reduced resolution we have to first upsample and then unskew rows back to original position. Upsampling is done using bicubic interpolation. Special care is taken at the boundary of masks in order to avoid issues with fractional data availability and oscillation artifacts in the reconstruction of high resolution data near mask boundaries. The error introduced by resampling, measured by $L^1$ loss function (\ref{cost}) and (\ref{distance}) on the whole dataset, is $0.012$, which is much smaller than the mean values of $L^1$ validation loss function obtained during model training, which is $0.020$. This provides evidence that the error introduced by computation on downsampled data is less significant in comparison to the inherent error introduced from our models.

\paragraph{Convolutional autoencoder hyperparameters.}
The described model is implemented in Python using the PyTorch library \citep{Paszke2019}. For the encoder part we used from $2$ to $11$ layers, so after the first layer which changes the number of channels from $d_{in}$ to $d_{ch}$ we have $1$ to $10$ additional outer, reducing and inner layers. So $1\leq N_{outer}+N_{reduce}+N_{inner}\leq 10$, where we prescribe $N_{outer}\geq 1$ and $N_{reduce}\leq 5$. Taking into account the decoder as explained in the previous sections, the total number of layers in a model is between $5$ and $23$. We include an additional so-called dropout layer between the encoder and decoder parts which reduces overfitting during training \citep{Nitish2014}. The dropout layer is a function having one hyperparameter $p\in[0,1]$ called the dropout percentage. The dropout layer is an identity map from $\mathbb{R}^k$ to itself which is multiplied element-wise by a random vector $\mathbf{v}\in\{0,1\}^k$. Every coordinate in $\mathbf{v}$ is sampled independently, being $1$ with probability $1-p$ and $0$ with probability $p$, every time the dropout layer is evaluated. In our case $k=d_\mathrm{ch} \times \left(W/2^{N_\mathrm{reduce}}\right) \times \left(H/2^{N_\mathrm{reduce}}\right)$.

For the number of channels we take $d_{ch}=64$ and the convolution kernel size we fix at $5\times 5$. In every layer except last we use the SELU nonlinear function \citep{Klambauer2017}. Also, each convolutional layer except last includes batch normalization as we find it improves training convergence and generalization to unseen examples \citep{Ioffe2015}.

Depending on the number of layers, our model has approximately $2.3\cdot 10^5$ to $2.1\cdot 10^6$ parameters. Dimension of the latent space however is solely regulated by the number of reducing layers $N_{reduce}$ and amounts to $256\cdot 4^{5-N_{reduce}}$, which ranges from $256$ to $262144$. Notice that the Red Sea downsampled spatial dimensionality of input data is around $1855$, so the dimensionality of the resulting latent space can be smaller as well as larger than the downsampled input data. The number of temperature anomaly data points available for training is daily data matrix size multiplied by the number of days and the average percentage of available data which equals around $1.4\cdot 10^7$. This is one to two orders of magnitude larger than the number of model parameters, so we are confident our models do not overfit.

\paragraph{Input dimensionality.}
Using multi-day input for training the model to predict the day in the middle of the input block allows the model to utilize the rate of change in time and learn correlations in short time scales. Our models are provided $N_\mathrm{days}=1,3,11$ consecutive days at input. We further find that providing masks as input in addition to anomaly data improves training. Additionally, positional encoding may be concatenated to input and adds two channels.

\paragraph{Loss function and norm.}
Seeing as the marginal distribution of the provided real-world data is approximately Gaussian, it would be reasonable to use the $L^2$ norm as the optimization goal in (\ref{cost}) because it should minimize deviation of predictions with regard to actual values. Intriguingly, in our experiments we find almost no difference in the $L^1$ validation loss when $L^1$ cost function was used as opposed to the $L^1$ validation loss when the $L^2$ cost function was optimized. We conservatively decided to use the $L^1$ cost function to obtain possibly larger reconstruction errors but also capture a larger variance of extreme values.

\paragraph{Optimizer hyperparameters.}
Models are trained using the fast.ai library \citep{Howard2018}. For the optimizer we employ the Ranger algorithm \citep{Wright2019} which stabilizes the start of the training process using Rectified Adam (RAdam) optimizer \citep{Liu2020}. Additionally, to stabilize the rest of the training, parameter lookahead avoids overshooting good local minima in parameter space \citep{Zhang2019}. Flat-cosine one-cycle policy for learning rate and weight decay ensures convergence to a broad optimum which allows the trained model to generalize well \citep{Smith2018}. The following RAdam hyperparameters in particular influence convergence and generalizability of the trained model: maximum learning rate, weight decay factor, exponential decay rates of the first and second moments, number of epochs per training and training batch size. We fixed maximum learning rate to $0.003$, the number of epochs to $50$, and the RAdam exponential decay rates to $(0.95,0.999)$. Additionally, we regard dropout percentage as an optimizer hyperparameter.

To suitably tune weight decay, batch size and dropout hyperparameters, we use the following grid search algorithm. In order to avoid over- or under-fitting, for each set of hyperparameters a model is trained through several iterations of 50 epochs until we achieve a satisfactory ratio between validation and training losses between $1$ and $1.05$. We start the first iteration of model training with dropout percentage $0$, weight decay $0.3$ and batch size $32$. In each iteration, we either decrease or increase regularization depending on whether the ratio of validation and training loss is less than $1$ or greater than $1.05$. To decrease regularization we decrease dropout percentage, weight decay, or batch size. Notice that since we use batch normalization in each convolutional layer, slightly decreasing batch size seems to actually decrease and not increase regularization contrary to expectation. To increase regularization we increase dropout percentage or weight decay. If in one iteration step we reach the ratio of validation and training loss less than $1$, and in the next step it is greater than $1.05$, or vice versa, for the following iteration we use a weighted linear interpolation of dropout percentage and weight decay hyperparameters from previous two iterations. At the end of this algorithm we select the model iteration with the the lowest validation loss using the missing data mask \( m^{(n)}_\mathrm{orig} -  m^{\prime(n)}\) as a relevant metric for the quality of reconstruction. We consider such a model to be well-trained.

\paragraph{Prediction ensembling.} One of the most important aspects that we use in our method is independent training of an ensemble of models with different convolutional autoencoder hyperparameters. A number of predictions for missing temperature anomaly values are inferred from each of the well-trained models. Afterwards, all of those predictions are ensembled to calculate the empirical distribution of wanted temperature anomaly extremes.

\section{Results and Discussion}
\label{res_and_dis}
For the solution of Data Challenge problem we ensemble predictions made by a set of $155$ models with different combinations of convolutional autoencoder hyperparameters that satisfy $1\leq N_{outer}+N_{reduce}+N_{inner}\leq 10$, with $N_{outer}\geq 1$, $0\leq N_{reduce}\leq 5$ and $N_{inner}\geq 0$. Each model was trained independently using the described grid search to select optimizer hyperparameters that produce well-trained models. The slowest observed training took $13$ iterations. On average it took $3.27$ iterations to reach a well-trained model, for a cumulative of $507$ trained models. For only one out of $155$ final models, the algorithm failed to achieve the targeted validation and training loss ratio which ended at only $1.09$.

Measured on a quad-core computer system with Nvidia RTX 2070 8 GB RAM GPU, the worst-case training time per model and per one iteration of hyperparameter grid search was around $1$ hour and $20$ minutes. The average training time was around 43 minutes. Cumulative training time for the full ensemble of $155$ well-trained models was approximately $15$ days.

For every of $155$ trained models, $20$ full historical predictions were inferred with a total of $3100$ complete spatio-temporal reconstructions of Red Sea temperature anomalies. These were used in place of missing data to calculate the empirical distribution of temperature anomaly extremes over space and time at locations specified by the Data Challenge problem. Timed on the above equipment, this operation took about $20$ minutes per model and finished in about $2$ days for all $155$ models.

We made altogether seven different runs of $155$ models ensemble training and inference, while varying $N_{\mathrm{days}}=1,3,11$ and training with or without positional encoding. This alone accounts for six different combinations of hyperparameters, while the seventh run was again using $N_{\mathrm{days}}=3$ and with positional encoding. The best $\overline{\mathrm{twCRPS}}$ score achieved was $3.581$ for $N_{\mathrm{days}}=3$ and without using positional encoding.

As a contrast, in our original second place solution to the Extreme Value Analysis $2019$ Data Challenge we also used an ensemble of predictions, but it used only $8$ models. These models were all trained and evaluated with similar convolutional autoencoder hyperparameters but on full-resolution data without resampling, using single-day input and no positional encoding. They were trained with the Adam optimizer \citep{Kingma2017} without hyperparameter tuning, and at that time we were using both $L^1$ and $L^2$ norms for model training. Then we reached the $\overline{\mathrm{twCRPS}}$ score of $4.667\cdot 10^{-4}$, so our current result is a significant improvement.

\begin{figure}
	\includegraphics[clip,width=1.0\linewidth]{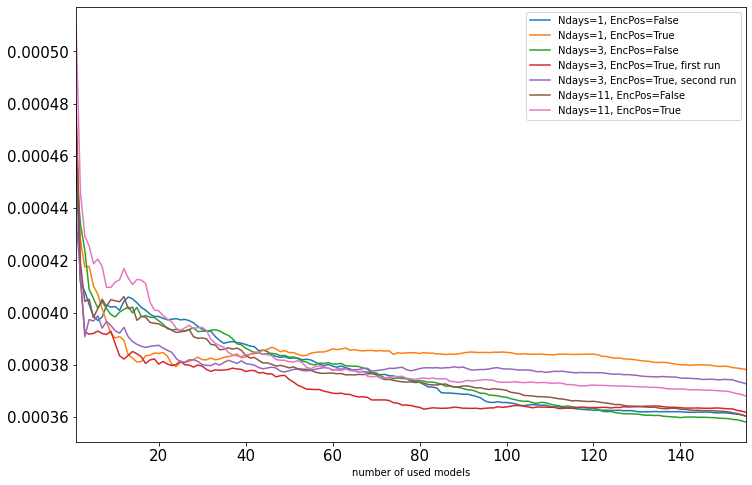}
	\caption{Graph of the $\overline{\mathrm{twCRPS}}$ score of a sub-ensemble prediction depending on the number of models used to form a sub-ensemble. Seven ensembles from Table \ref{table:scores} are used. The models are sorted ascending by validation loss evaluated at additionally damaged data. The first $M$ models are chosen for each sub-ensemble.}
	\label{fig:score2}
\end{figure}

Although every run is expensive in computer time, we decided to separately train two ensembles of models with the same model hyperparameters to assess the variability in score introduced by stochasticity in our model training procedure. In the end we calculate the $\overline{\mathrm{twCRPS}}$ score for every ensemble and present the results in Table \ref{table:scores} and Figure \ref{fig:score2}.  By looking at this data, we could hypothesize that positional encoding is making $\overline{\mathrm{twCRPS}}$ score worse. However, it is important to notice the relatively large variability in score between runs $4$ and $5$ which have the same model hyperparameters. Unfortunately, it seems that the difference we see between different runs can largely be attributed to stochasticity in model training procedure.

\begin{table}[htb]
	\caption{$\overline{\mathrm{twCRPS}}$ scores computed for seven training runs of $155$ model ensembles using different value combinations of $N_{\mathrm{days}}$ and positional encoding hyperparameters. Number of channels $d_{ch}=64$ in every ensemble. Notice the variation of score for runs $4$ and $5$ which share the same values for $N_{\mathrm{days}}$ and positional encoding (see text).}
	\begin{center}
		\begin{tabular}{|l|l||c|c|c|}
			\hline
			run no. & $N_{\mathrm{days}}$ & pos.\ enc. & $\overline{\mathrm{twCRPS}} \ / 10^{-4}$ \\
			\hline
			$1$ & $1$ & No & $3.604$  \\
			\hline
			$2$ & $1$ & Yes & $3.782$  \\
			\hline
			$3$ & $3$ & No & $3.581$  \\
			\hline
			$4$ (first) & $3$ & Yes & $3.618$  \\
			\hline
			$5$ (second) & $3$ & Yes & $3.728$  \\
			\hline
			$6$ & $11$ & No & $3.603$  \\
			\hline
			$7$ & $11$ & Yes & $3.681$  \\
			\hline
			
		\end{tabular}
	\end{center}
	\label{table:scores}
\end{table}	

We are also interested in what happens if we try to take a smaller ensemble of predictions. The ensemble produced by $155$ models takes quite a long time to train and evaluate so it would be of benefit if a smaller ensemble had comparable performance regarding the $\overline{\mathrm{twCRPS}}$ score.

First let us consider only a trivial ensemble created by a single model. Let us take the model with the lowest validation loss out of all $155$ models when evaluated at additionally damaged data \(m^{(n)}_\mathrm{orig} -  m^{\prime(n)}\), i.e., the best model from ensemble in run no.\ $4$ in Table \ref{table:scores}. Let us sample from this model full $3100$ data reconstructions, the same number of samples as the previously discussed large ensembles. We get the modest $\overline{\mathrm{twCRPS}}$ score of $4.804\cdot 10^{-4}$, which means that our large ensemble indeed helps improve the prediction quality over a single best-performing model. Notice that taking $3100$ samples for a single model does not significantly improve the score when compared to only $20$ samples ($4.802\cdot 10^{-4}$).

\begin{figure}
	\includegraphics[clip,width=1.0\linewidth]{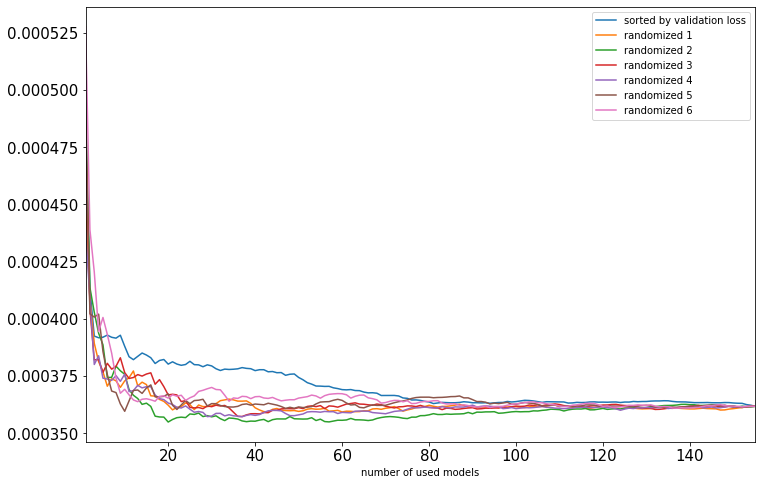}
	\caption{Graph of the $\overline{\mathrm{twCRPS}}$ score of a sub-ensemble prediction depending on the number of models used to form a sub-ensemble. The ensemble was trained using $N_{\mathrm{days}}=3$ and positional encoding. The models are either sorted ascending by validation loss evaluated at additionally damaged data or the order is randomized. The first $M$ models are chosen for each sub-ensemble.}
	\label{fig:score}
\end{figure}

Next, let us consider a family of ensembles, each ensemble a sub-ensemble consisting of the first $M$ out of $155$ well-trained models, ordered by the ascending validation loss evaluated at additionally damaged data \(m^{(n)}_\mathrm{orig} -  m^{\prime(n)}\). For $M=1$ we get the trivial ensemble already considered, and for $M=155$ we get the full ensemble. For comparison, let us take a couple of randomized model orders of run no. $4$ and produce the same sub-ensembles, by taking only the first $M$ out of $155$ models, see Figure \ref{fig:score}. Random picking the sub-ensemble wins over ordered picking, but we still have to put quite a large number of models into our ensemble to get close to the $\overline{\mathrm{twCRPS}}$ score achieved using all of $155$ models.

Alternatively, in Figure \ref{fig:score2} we consider taking only the first $M$ out of $155$ models for each of the seven trained model ensembles. Ordering of the models is again produced using an ascending sort by validation loss evaluated at additionally damaged data.

The grid search algorithm that tunes optimizer hyperparameters clearly helps with the quality of prediction. To illustrate this we took $155$ models trained for 50 epochs once, of run no. $4$, using fixed optimizer hyperparameters without any additional tuning. Many of these models were over- or under-trained. We calculated the $\overline{\mathrm{twCRPS}}$ score to be $3.970\cdot 10^{-4}$, meaning that proper hyperparameter selection and well-trained models improved the score by about $10\%$.

Training a large ensemble of models is prohibitively expensive regarding computation time and resources. Therefore it is difficult to investigate the impact of each decision in selecting individual model hyperparameters. It is far from clear which hyperparameters (number of layers, kernel size, number of channels, number of days at input...) had the greatest impact on the improvement of our prediction. A detailed ablation study with a full-blown set of models unfortunately may well require months or years to train and evaluate using our currently available hardware.

Figure \ref{fig:score} indicates that a randomly chosen ensemble of only about $25$ out of $155$ current models could prove sufficient for an ablation study. We have indeed made first attempts in that direction by selecting $5$ different models as templates and varying $d_{ch}\in\{64,128,256\}$, $N_{days}\in\{1,3,11\}$ as well as whether positional encoding is used. This resulted in $90$ well-trained models partitioned in $18$ different small ensembles. Despite each small ensemble containing only 5 different models, there are preliminary indications that positional encoding is very helpful and that a large block of $N_{days}=11$ at the input performs better than $N_{days}\in\{1,3\}$, although it seems that it happens only in cases when $d_{ch}\in\{128,256\}$. We hypothesize that the relatively small number of channels, $d_{ch}=64$, is the main limiting factor for models with large $N_{days}$ and positional encoding which otherwise might be able to show their strengths and significantly decrease the score, see Table \ref{table:scores}. We also hypothesize that the other most probable suspect that hampers further reduction in score is the spatial down-sampling of data. Namely, in our case it reduces the amount of available training data by a factor of $9$ and introduces additional prediction error in the up-sampling step. Unfortunately, either increasing $d_{ch}$ or working with full resolution models is extremely expensive not just regarding computation time, but also regarding GPU memory used, which makes it unfeasible for our currently available computer system.

The presented technique relies on a suitable choice of masks that describe additional damage. In domains where such data loss is easily generated, such as tabular data or low-dimensional time series, our technique could prove to be useful. For image-like datasets found in medicine, geology, climatology, etc., an extensive study is needed to assess the influence of added damage and the distribution of imputed noise on the quality of recovered data.

We also note there are other viable model architectures that ought to be explored. For instance, in this work we use the simplest convolutional layers for our autoencoder. Instead, better generalization might be obtained using ResNet \citep{He2015} or U-Net architectures which use skip connections as high-resolution pathways between distant layers \citep{Ronneberger2015}. In particular, U-Net places skip connections between corresponding encoder and decoder layers to preserve fine detail. Even though it is originally used for medical image segmentation or classification, U-Net might prove to be a good fit for regression problems such as ours as it is e.g., used to model MisGAN's generators \citep{Li2019}. Going further, in our dataset possibly the largest source of untapped information lie in long-term temporal correlations which we currently underutilize. A more extensive study of time-domain information is needed. Here, image latent space could be used as input to dedicated recurrent neural networks or even novel attention-based models currently explored by the natural language processing community \citep{Vaswani2017}.

\section{Concluding remarks}
In this work we present a solution to the Extreme Value Analysis 2019 Data Challenge. A technique is described to recover missing data by training an ensemble of models on additional data damage we introduce ourselves. Sampling from autoencoder-like approximations of observed data distributions provides a feasible way to analyze complex dynamics of geophysical phenomena. The described approach seems amenable to be applied in other areas of basic and applied research with rare and extreme events as well as complement existing extreme value theory techniques.

%\begin{figure}
%  \includegraphics{example.eps}
%\end{figure}
%\begin{figure*}
%  \includegraphics[width=0.75\textwidth]{example.eps}
%\end{figure*}
%\begin{table}
%\begin{tabular}{lll}
%\hline\noalign{\smallskip}
%\end{tabular}
%\end{table}

\begin{acknowledgements}
We thank Ivan Balog for enlightening discussions.
\end{acknowledgements}

\section*{Conflict of interest}
The authors declare that they have no conflict of interest.

\bibliographystyle{spbasic}      % basic style, author-year citations
\bibliography{blackbox}   % name your BibTeX data base

\end{document}